\documentclass[letterpaper, 10 pt, conference]{ieeeconf}  

\IEEEoverridecommandlockouts                              

\usepackage{color}
\usepackage{caption}
\usepackage{graphics} 
\usepackage{amsmath} 
\usepackage{amssymb}  
\usepackage{biblatex} 
\usepackage{booktabs}
\usepackage{graphicx}
\usepackage[capitalise]{cleveref}
\let\labelindent\relax
\usepackage{enumitem}
\usepackage{tablefootnote}

\usepackage[textsize=tiny]{todonotes}

\title{\LARGE \bf
N-QR: Natural Quick Response Codes\\for Multi-Robot Instance Correspondence
}

\author{Nathaniel Moore Glaser$^{1,2}$, Rajashree Ravi$^{2}$, and Zsolt Kira$^{1}$\\ 
${^1}$Georgia Institute of Technology\\
${^2}$Bowery Farming\\
\texttt{$\{$nglaser,zkira$\}$@gatech.edu}
}
\begin{document}

\maketitle
\thispagestyle{empty}
\pagestyle{empty}

\begin{abstract}
Image correspondence serves as the backbone for many tasks in robotics, such as visual fusion, localization, and mapping.
However, existing correspondence methods do not scale to large multi-robot systems, and they struggle when image features are weak, ambiguous, or evolving.
In response, we propose \textit{Natural Quick Response} codes, or \textit{N-QR}, which enables \textit{rapid} and \textit{reliable} correspondence between large-scale teams of heterogeneous robots. 
Our method works like a QR code, using keypoint-based alignment, rapid encoding, and error correction via ensembles of image patches of natural patterns.  
We deploy our algorithm in a production-scale robotic farm, where groups of growing plants must be matched across many robots.  
We demonstrate superior performance compared to several baselines, obtaining a retrieval accuracy of 88.2\%.  
Our method generalizes to a farm with 100 robots, achieving a 12.5x reduction in bandwidth and a 20.5x speedup.  
We leverage our method to correspond 700k plants and confirm a link between a robotic seeding policy and germination.
\end{abstract}

\section{Introduction}
Many robotic tasks, such as visual localization and mapping, rely on matching the same features across image views.  
This process, commonly referred to as image correspondence, often requires prominent, static features to perform the matching (e.g. the rigid corners of buildings).
However, these types of features are not guaranteed, especially as robots venture into environments that have non-rigid features.

In this paper, we address one such environment---robotic agriculture.  In our setting, plants are grown by moving them through a sequence of specialized robot stations, in a process similar to a factory assembly line.
Ultimately, we seek to use the cameras of each robot station to track every single plant throughout its lifecycle.  By monitoring individual plants from seed to harvest, farmers can make key decisions about adjusting seeding patterns, water, light, and nutrients.

However, due to system limitations, our plants are shuffled between stations, and plant-level tracking must be performed without external markers (e.g. QR tags).  Instead, we must rely on the plants themselves as the unique identifiers.  This constraint makes tracking challenging, as plants are non-rigid, growing objects with ambiguous features.

To address these challenges, we propose \textbf{Natural Quick Response}, a learned approach for performing high-volume, ambiguity-prone correspondence between bandwidth-limited robots.  
\textbf{N-QR} aligns a candidate object to a uniform representation where it then ensembles and encodes image patches into compact, robust features for cross-robot comparison.

Our approach expands the operational domain of image correspondence along three dimensions: (1) \textbf{\textit{multi-robot scale}}, (2) \textbf{\textit{viewpoint heterogeneity}}, and (3) \textbf{\textit{visual ambiguity}}.  First, our algorithm scales to a farm that has thousands of communicating robots, each with their own sensors, actuators, and compute.  Second, it performs matching despite significant \textbf{\textit{visual dissimilarity}} between (a) cameras of different resolutions, lighting, and positioning and (b) objects that have changing visual features.
Third, it performs matching despite misleading \textbf{\textit{visual similarity}}, as the subjects that we are imaging (i.e. a grid of plants) have strong, ambiguous features, but weaker unique features.  

\begin{figure}
\vspace{2mm}
\centering
\setlength{\belowcaptionskip}{-15pt}
\includegraphics[width=\linewidth]{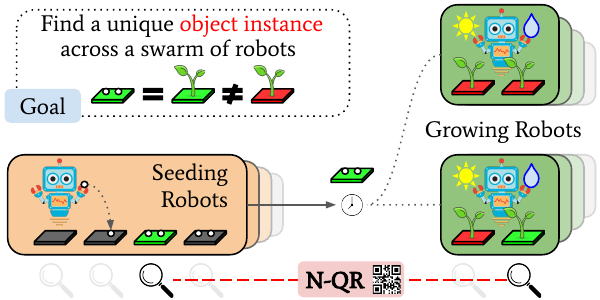}
\caption{\textbf{N-QR} uses naturally-occurring patterns and rapid encoding to help large teams of robots find a unique object.  This correspondence allows robotic farmers to track the same plant, from seed to harvest, without extra tagging hardware.
}
\label{fig:natural_qr_code}%
\end{figure} 

\begin{figure*}
\vspace{2mm}
\centering
\setlength{\belowcaptionskip}{-10pt}
\includegraphics[width=\linewidth]{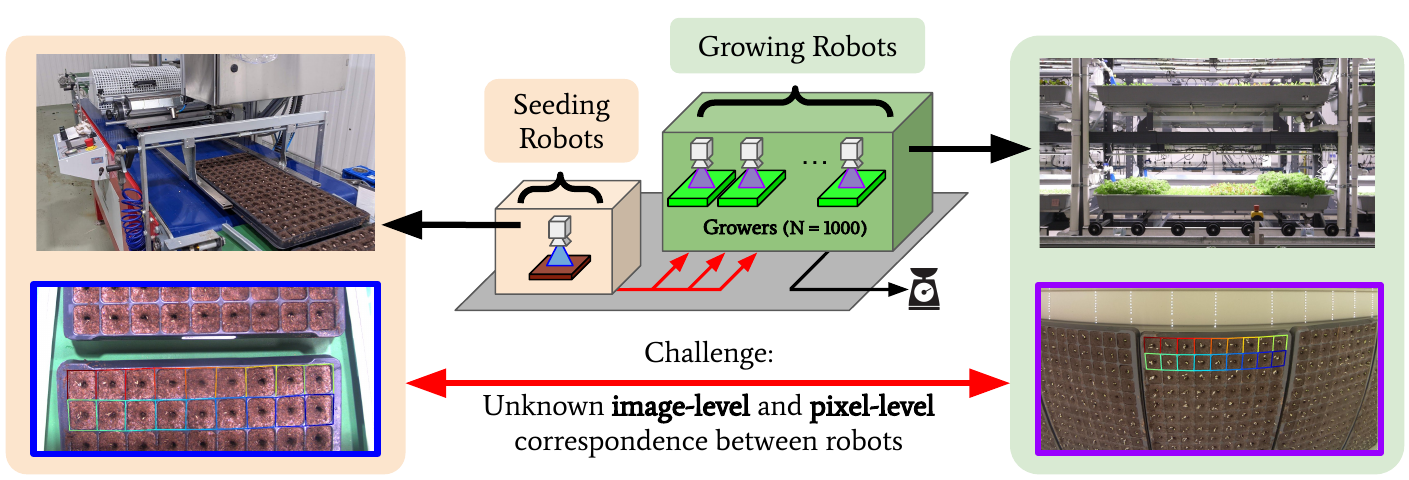}
\caption{\textbf{Robotic farming system and image matching challenge.}  A seeding robot (left) continuously plants seeds into ``rafts'' on a moving conveyor belt. After a germination period, these rafts are transferred to a stationary growing robot (right), who supports and monitors growth for the remaining duration of the plant lifecycle.  \textbf{\textit{Image-level, raft-level, and pixel-level correspondences are unknown between the seeding and growing robots.}}}
\label{fig:system_overview}%
\end{figure*} 

We summarize the contributions of our paper as follows:
\begin{itemize}
    \item We address the task of \textbf{\textit{large-scale}}, \textbf{\textit{multi-robot}} \textbf{instance correspondence} in an unprecedented research setting---a \textbf{production robotic farm} with \textit{thousands} of robots.
    \item Our method, \textbf{N-QR}, achieves a state-of-the-art image retrieval accuracy of \textit{\textbf{88.2\%}} via a novel, multi-tiered ensembling scheme.  This approach matches the same physical object \textit{despite drastic appearance changes}.
    \item Our bandwidth-efficient transmission policy allows each robot to iteratively describe its observations via a scheduled transmission of low-dimensional embeddings.  We leverage decentralized compute and reduce bandwidth by \textit{\textbf{12.5x}} and computation latency by \textit{\textbf{20.5x}}.
    \item Finally, we deploy this matching pipeline for multi-view agricultural insights.  Our method finds a link between our robotic seeding policy and resultant plant growth.
\end{itemize}

\section{Related Work}
Several domains relate to the task of \textbf{multi-robot instance correspondence} and \textbf{multi-view growth analysis}.  

\textbf{Image correspondence} methods~\cite{orb,lowe1999sift,muja2015fast,fischler1981ransac,sarlin2020superglue,lindenberger2023lightglue,truong2021pdc,truong2020gocor,glaser2021enhancing,glaser2021overcoming} seek to match features or perform dense alignments between two images, on a pixel level.
\textbf{Traditional sparse correspondence} methods~\cite{orb,lowe1999sift,muja2015fast,fischler1981ransac} rely on strong corners and geometrically-consistent features to compute and confirm matching keypoints between different images.  
\textbf{Learned sparse correspondence} methods~\cite{sarlin2020superglue,lindenberger2023lightglue} leverage learned feature descriptors, semantics, and relationships to match across broader featureless regions.  \textbf{Dense correspondence} methods~\cite{truong2021pdc,truong2020gocor} compute a dense pixel warping grid between images.  \textbf{Multi-robot dense correspondence}~\cite{glaser2021enhancing,glaser2021overcoming} methods address the added challenge of corresponding across several agents, often while paying heed to bandwidth and computational constraints.  However, these methods completely fail in our setting (see Sec.~\ref{sec:dense_matching}).  In response to these failures, we decompose the instance correspondence problem into two subproblems: (1) aligning the pair of images to a normalized representation and (2) performing discrete image (block) matching.

\textbf{Keypoint detection} works~\cite{law2018cornernet,he2017mask} are useful for our alignment problem.  These works use CNN architectures to generate a heatmap mask from which keypoins are extracted.  We leverage similar techniques to detect keypoints within our scene, such as seeding tray corners and vertices, which we then use for image warping.

\textbf{Discrete image matching} methods~\cite{schroff2015cvpr,koch2015siamese} seek to find matches on an image level.  
\textbf{Metric learning}~\cite{schroff2015cvpr} approaches learn image-level descriptors such that metric distances between similar images are low compared to distances between dissimilar images.  One popular instance of this approach is the \textbf{Siamese Network}~\cite{koch2015siamese}, which uses a shared neural network encoder to produce these image-level descriptors.  Our work similarly uses a metric learning objective.  However, unlike these prior works, our approach leverages multiple tiers of patch ensembling to overcome noisy and misleading inputs.

\textbf{Content Based Instance Retrieval (CBIR)} methods~\cite{chen2022deep,Weyand_2020_CVPR,zhao2017spatial,song2017deep,song2018binary,muller1999efficient} aim to improve accuracy by extracting distinctive features and reducing the impact of image clutter when efficiently querying a large database. While our approach addresses similar challenges, our dataset presents a higher level of complexity, characterized by minimal scene variation and a notable degree of visual similarity among instances, distinguishing it from commonly used datasets such as \textbf{GLDv2}~\cite{Weyand_2020_CVPR}. Moreover, previous works have focused on addressing search efficiency concerns using methods like \textbf{deep hashing}~\cite{zhao2017spatial,song2017deep,song2018binary} and inverted files~\cite{muller1999efficient}. Our method employs a bandwidth efficient iterative transmission policy with increasing feature sizes to minimize the total number of packets required for accurate matching. 

\textbf{Multiple instance learning} looks at multiple instances to determine an overall classification.  Several works~\cite{li2021deep,ilse2018deepmil} consider multiple image patches of a cancer cell before rendering a final verdict, which is especially useful when individual patches are noisy or misleading.  Our approach extends this idea from a classification setting to a metric learning setting, especially for robust image matching.

\textbf{Yield estimation} methods use plant phenotypes~\cite{mokhtar2022using} or overhead camera information~\cite{jung2015lettuce,zhang2020growth} to predict the final harvest mass of a crop.
Similarly, we evaluate crop yield in our system by calculating leaf area from overhead camera images, a metric strongly correlated with harvest yield, as demonstrated in prior studies.

Unlike prior research~\cite{riera2021deep}, which performs \textbf{multi-view yield estimation} by capturing the same plant from different angles, our approach employs multiple views differently. We gather complementary information from heterogeneous sensors observing various stages of plant growth, thus enriching our analytical insights.
\section{Methodology}
\begin{figure}
\vspace{2mm}
\setlength{\belowcaptionskip}{-15pt}
\centering
\includegraphics[width=1.0\linewidth]{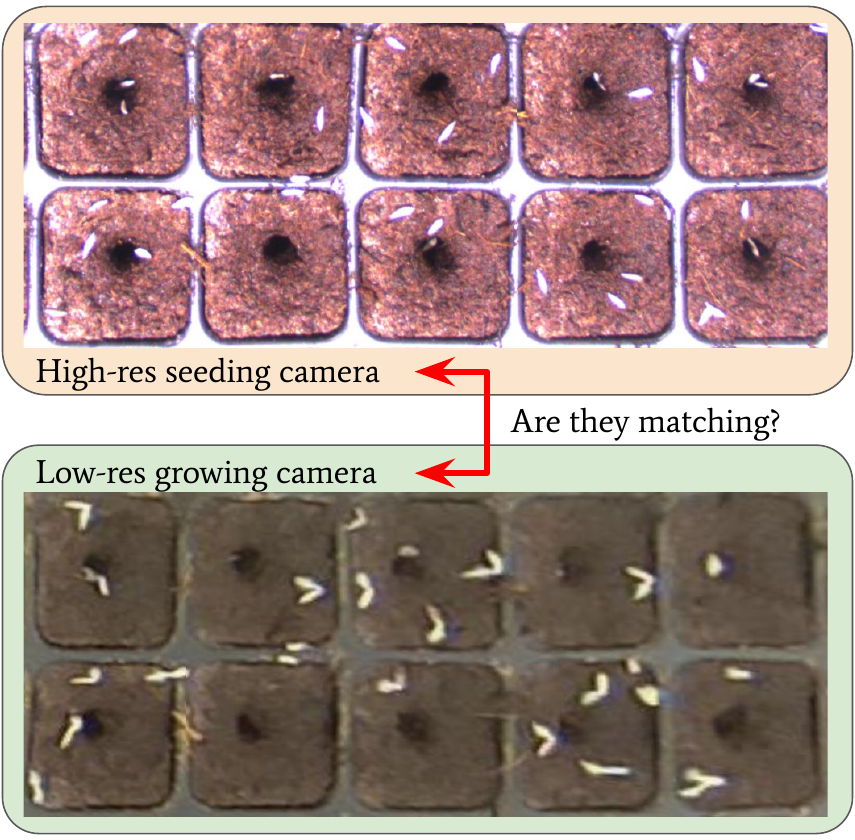}
\caption{\textbf{Matching difficulty}. These two normalized raft images ($\mathbf{R}$ and $\mathbf{R^{\prime}}$) constitute a positive match.  As an exercise for the reader, we encourage you to find the features that support and discourage this match.}
\label{fig:matching_example}%
\end{figure} 

\subsection{System}
As depicted in fig.~\ref{fig:system_overview}, our system is a production-scale \textbf{vertical farm}\footnote{\textbf{Vertical Farm:} A ``small'' footprint, indoor farm that grows crops in a space-efficient, stacked configuration}.
Within this system, a set of agricultural robots $\{\mathbf{S}_0, \mathbf{S}_1, \mathbf{G}_0, ..., \mathbf{G}_{3000}\}$ work together to grow a plant through different parts of its extended lifecycle. 

\textbf{Seeding:} Each seeding robot $\mathbf{S}_i$ sits above a conveyor belt, where it drops seeds into a sequence of $M$ rafts\footnote{\textbf{Raft:} A grid of dirt that can be irrigated by flooding it with water}.  For each raft $m$, the robot captures a top-down image $\mathbf{X}_i^m$, containing one \textbf{normalized raft image} $\mathbf{R}_i^m$:
\begin{equation}\label{eq:raft_warping_seed}
\mathbf{R}_i^m = \mathbb{W}_i^m(\mathbf{X}_i^m),
\end{equation}
where $\mathbb{W}$ is an warping function that extracts a cropped, uniform grid of dirt cells $\mathbf{R}$ from $\mathbf{X}$, as in figs.~\ref{fig:system_overview} and ~\ref{fig:matching_example}.  

\textbf{Germination:} Next, the rafts are moved into a chamber for germination \textbf{and are not tracked during this period}\footnote{\textbf{Raft Tracking:} To enable raft-level tracking, we would need to retool and retrain operators.  Rather than overhaul the farm with QR codes and scanners, this work explores the minimally-invasive question: \textbf{can we use the raft itself as a QR code?}}.

\textbf{Growth:} After germination, the rafts are manually placed onto benches\footnote{\textbf{Bench:} An open container used to hold, move, and irrigate several rafts}.   A robot then moves each bench to a designated growing robot $\mathbf{G_j}$.  This robot $\mathbf{G_j}$ supplies light and water to the plants for the remainder of the lifecycle of the plant.  At regular intervals, each growing robot captures a top down image $\mathbf{Y}_j^t$, which contains 10 normalized raft images:
\begin{equation}\label{eq:raft_warping_grow}
\mathbf{\bar{R}}_{j,q}^t = \mathbb{W}_{j,q}^t(\mathbf{Y}_j^t), \text{ for } q=\{0,...,9\}.
\end{equation}

It is important to note that each normalized raft image $\mathbf{\bar{R}}$ captured by a growing robot is a time-delayed version of a raft image $\mathbf{R}$ captured during seeding:
\begin{equation}\label{eq:raft_growth}
\mathbf{\bar{R}} = \mathbb{F}(\mathbf{R}, t, s, e),
\end{equation}
where the general farming process $\mathbb{F}$ induces visual changes in $\mathbf{R}$ based on the time since seeding $t$, seeding configuration $s$, and environmental influences $e$ such as lighting and water.  

Oftentimes, the visual changes produced by $\mathbb{F}$ are so severe that traditional dense matching pipelines fail~\cite{orb,lindenberger2023lightglue,truong2021pdc}.  These changes include the following: (1) object geometry changes (since the seeds have germinated and grown), (2) minor to major positional changes (since the plants may be jostled between stages), and (3) illumination and resolution changes.  
A matching example of $\mathbf{R}$ and $\mathbf{\bar{R}}$ is shown in fig.~\ref{fig:matching_example}.

\subsection{\textbf{Task:} Multi-Robot Dense Matching}
Given this farming system, we first address the task of \textbf{instance correspondence} between the raft in $\mathbf{X}$ and $\mathbf{Y}$.  After substituting eq.~\ref{eq:raft_warping_seed} and eq.~\ref{eq:raft_warping_grow} into eq.~\ref{eq:raft_growth} and setting $t=10$ (the earliest available growing robot image), we yield an equation that summarizes our challenge:
\begin{equation}\label{eq:pixel_matching_simple}
\mathbb{W}_{j,q}(\mathbf{Y}_j) = \mathbb{F}(\mathbb{W}_i^m(\mathbf{X}_i^m)).
\end{equation}
Namely, we seek to find object pixels in $\mathbf{X}_i^m$ that map to object pixels in $\mathbf{Y}_j$, despite drastic appearance changes (induced by $\mathbb{F}$), unknown robot association ($i$, $j$), multiple candidates per image ($q$), and unknown sub-image alignment ($\mathbb{W}_{j,q}$, $\mathbb{W}_i^m$).  
To tackle these challenges, we propose an \textbf{alignment} and \textbf{discrete matching} pipeline.  

\subsection{Alignment}
To compute the normalized raft images as described in eqs.~\ref{eq:raft_warping_seed} and \ref{eq:raft_warping_grow}, we perform the following:
\begin{enumerate}
    \item \textbf{Raft BBox NN\footnote{\textbf{BBox NN:} bounding box neural network}:} For each image, use a keypoint NN to identify $Q \times 2$ corner points for each raft, then crop the raft.  
    \item \textbf{Raft Vertex NN:} For each raft image, use a second keypoint NN to extract $H \times W$ grid vertices.  Group points to represent the four corners of each dirt cell.
    \item \textbf{Patch Extraction:} For each set of vertex corners, warp the source image into a square target image.  Recombine these cells to form a \textbf{normalized raft image}, as in fig.~\ref{fig:matching_example}.    
\end{enumerate}

\subsection{Discrete Matching}
The objective of the discrete matching pipeline is to satisfy eq.~\ref{eq:raft_growth}, given the discrete choices of $\{\mathbf{R}_i^m\}$ and $\{\mathbf{\bar{R}}_{j,q}\}$, as generated by the warping procedure:
\begin{equation}\label{eq:raft_growth_discrete}
\mathbf{\bar{R}}_{j,q} = \mathbb{F}(\mathbf{R}_i^m)
\end{equation}
In other words, we want to find the correct choice of $\hat{m}$, $\hat{i}$, $\hat{j}$, and $\hat{q}$ out of all potential choices $(|G| \times Q) \times (|S| \times M)$.
The total number of pairwise choices for the entire farm is approximately $10^6$.

To address these challenges, we propose a metric learning approach that uses a decentralized, bandwidth-efficient feature extractor to generate $\mathbb{F}$ invariant embeddings.  Namely, we propose two parallel neural networks $\mathbf{\Phi}_S$ and $\mathbf{\Phi}_G$ to compute embeddings $f$ and a pairwise distance $l_2$:
\begin{equation}\label{eq:feature_extraction}
\begin{split}
f_i^m = \mathbf{\Phi}_S(\mathbf{R}_i^m), \qquad \bar{f}_{j,q} = \mathbf{\Phi}_G(\mathbf{\bar{R}}_{j,q}) \\
l_2(\mathbf{{R}}, \mathbf{\bar{R}}) = {\lVert \bar{f}_{j,q} - f_i^m \rVert}_2
\end{split}
\end{equation}
To satisfy eq.~\ref{eq:raft_growth_discrete}, we want the distance for a positive match $l_2(\mathbf{{R}}, \mathbf{\bar{R}}^+)$ to be smaller than the distance of a negative match $l_2(\mathbf{{R}}, \mathbf{\bar{R}}^-)$ by some margin $\alpha$.  We adopt the triplet loss objective~\cite{schroff2015cvpr}:
\begin{equation}\label{eq:metric_learning}
\mathcal{L}(\mathbf{{R}}, \mathbf{\bar{R}}^+, \mathbf{\bar{R}}^-) = \text{max}(l_2(\mathbf{{R}}, \mathbf{\bar{R}}^+) - l_2(\mathbf{{R}}, \mathbf{\bar{R}}^-) + \alpha, 0)
\end{equation}

\begin{figure}
\vspace{2mm}
\setlength{\belowcaptionskip}{-15pt}
\centering
\includegraphics[width=\linewidth]{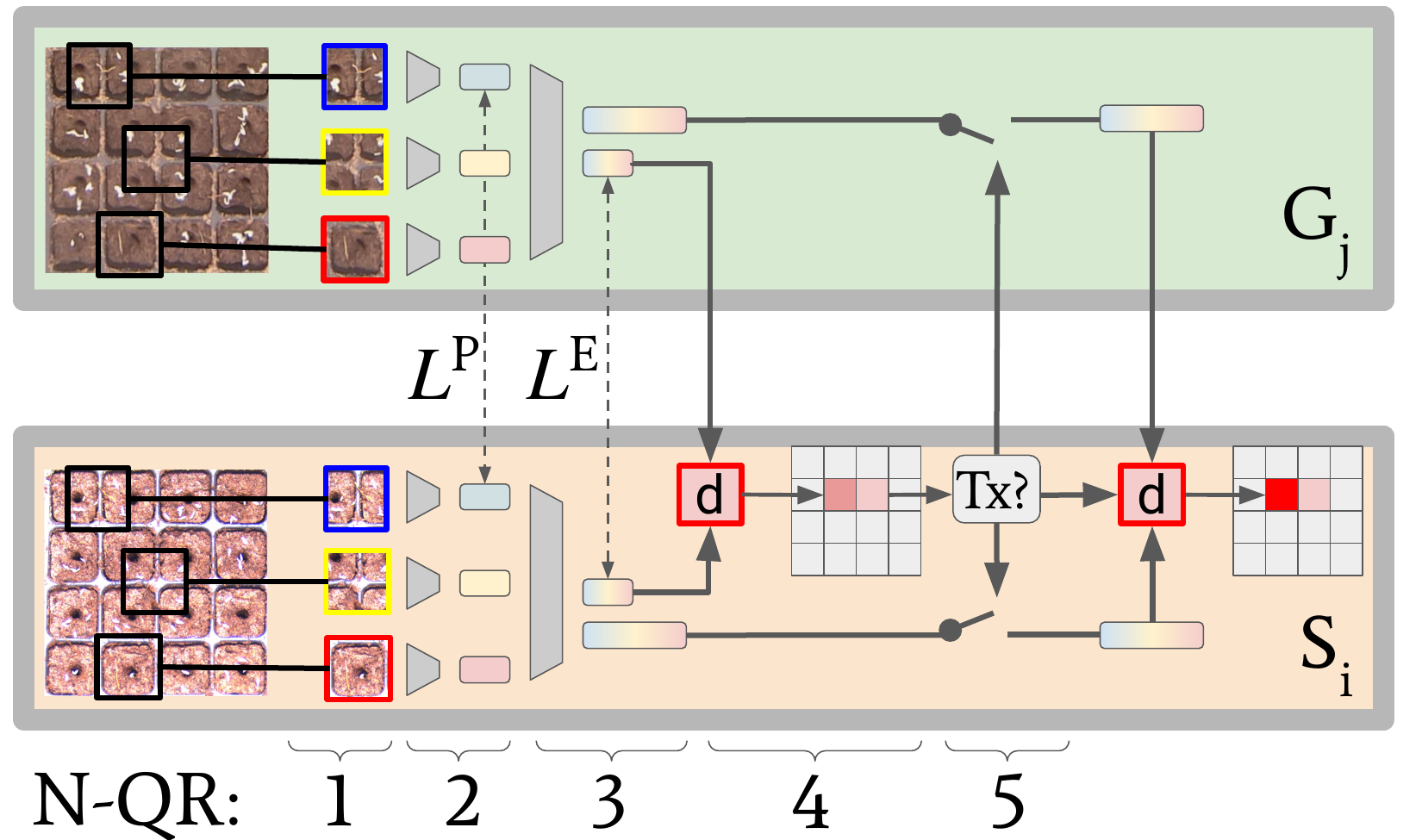}
\caption{\textbf{N-QR} iteratively transmits feature embeddings until a minimum distance threshold is met.
}
\label{fig:comparison_network_bw}%
\end{figure}

\textbf{1-vs-1 Raft Matching:}
In order to make $\mathbf{\Phi}_S$ and $\mathbf{\Phi}_G$ invariant to the extreme influences of $\mathbb{F}$, we propose a specialized comparison network based on image patch ensembling, as shown in fig.~\ref{fig:comparison_network_bw}.  Decomposing images into patches lets us consider partial raft images, apply an intermediate training signal at the patch level, and increase our dataset size (via random permutations and augmentations for each patch). To over come insufficient or misleading information in noisy patches, our method aggregates their features in an ensemble. 
Our method involves randomly sampling image patches from the same locations in $\mathbf{R}$ and $\mathbf{\bar{R}}$, extracting patch-level embeddings using a ResNet extractor, and stacking them along their channel dimension using a fully connected network to generate a final image-level embedding $f$ and $\bar{f}$. Triplet losses $\mathcal{L}^P$ and $\mathcal{L}^E$ are used to train patch-level and image-level embeddings, respectively.

Based on the training objective, the distance between the features of a matching raft should be smaller than that of a non-matching raft:
\begin{equation}\label{eq:feature_comparison}
{\lVert \bar{f}_{\hat{j},\hat{q}} - f_{\hat{i}}^{\hat{m}} \rVert}_2 + \alpha < {\lVert \bar{f}_{j,q} - f_i^m \rVert}_2
\end{equation}

\textbf{1-vs-Many Raft Matching:}
The previous procedure evaluates a single match candidate \textit{pair} $\mathbf{R}$ and $\mathbf{\bar{R}}$.  Beyond this capability, we also want to ``retrieve'' the right match among numerous incorrect ones. Ideally then, the smallest computed pairwise distance between $\mathbf{R}_i^m$ and all $\{\mathbf{\bar{R}}_{j,q}\}$ would correspond to the correct match out of $|G| \times Q$.

\textbf{Multi-Pass 1-vs-Many Raft Matching:}
To enhance accuracy over our base approach, we run our pairwise matching network with random patch samples in multiple passes and then average the pairwise distances across these batches.

\textbf{Decentralized Processing:}
Our system comprises of heterogeneous robots with varying compute capabilities: the many growing robots have cost-effective, weaker CPUs, while the relatively few seeding robots have desktop processors.
Instead of broadcasting all raw images to a centralized processor, we use the natural parallelism of our cluster of growing robots. Each robot performs its own warping and feature extraction, with $\mathbf{\Phi}_S$ and $\mathbf{\Phi}_G$ and features $\bar{f}_{j,q}$ are then broadcast to each seeding robot for $|G| \times Q$ pairwise comparisons. 

\textbf{Bandwidth Efficient Transmission Policy:}
We further conserve bandwidth via an iterative transmission policy, as shown in fig.~\ref{fig:comparison_network_bw}.  Since our system generates a significant amount of network chatter, we want to minimize the cumulative packet size required to perform accurate matching.  Therefore, we propose to iteratively broadcast denser and denser feature representations $\bar{f}_{\text{tx}}$, based on the distance between the smallest and second smallest pairwise distances $\delta = d_1 - d_0$ relative to margin $\alpha$:
\begin{equation}\label{eq:transmission}
\bar{f}_{\text{tx}}(\mathbf{\bar{R}}, \delta, c) = 
\begin{cases}
    \mathbf{\Phi}_G^{16}(\mathbf{\bar{R}}) ,& \text{if } \delta < \alpha \land c < 10 \\
    \mathbf{\Phi}_G^{128}(\mathbf{\bar{R}}) ,& \text{if } \delta < \alpha \land c \geq 10 \\
    \text{stop transmit}, & \text{if } \delta \geq \alpha \lor c \geq 20 \\
\end{cases}
\end{equation}
where $c$ is the index of the transmission and $\mathbf{\Phi}_G^{16}$, $\mathbf{\Phi}_G^{128}$ are networks that extract features at sizes of $16$ and $128$, respectively.  We incorporate this transmission policy into our ensembling scheme---whereby each transmitted feature is used to compute a pairwise distance, which we combine into a running average distance matrix.

\subsection{\textbf{Task:} Multi-View Seed-Growth Analysis}
Ideally, the seeding robot plants all seeds into the central hole within each dirt cell. This recessed area provides an ideal seed germination environment with its darkness and moisture.  In practice, however, seeds often stray from this desired location, yet they still manage to grow.  We seek to answer the question: how crucial is it for our robot to plant seeds in the hole?  Should farms invest in optimizing seed placement, or is the current system sufficient? We hypothesize that seeds in the hole have a higher germination rate based on intuition.

To test our hypothesis, we extract all grid cell patches $\mathbf{P}$ and $\mathbf{\bar{P}}$ for each raft image $\mathbf{\bar{R}}$ and $\mathbf{R}$. Next, we want to determine how a seeding pattern $s$ observed in a dirt cell $\mathbf{P}$ influences growth $h$ observed in 
$\mathbf{\bar{P}}^t$ over growing time $t$:
\begin{equation}\label{eq:raft_growth_conditional}
    s =  \mathbf{\Psi}(\mathbf{P}), \qquad h(t) =  \mathbb{H}(\{\mathbf{\bar{P}}^0,...,\mathbf{\bar{P}}^t\})
\end{equation}

where $\mathbf{\Psi}$ is a keypoint detection network used for predicting seed locations and $\mathbb{H}$ is a procedure for measuring growth over time.

\begin{figure}
\vspace{2mm}
\setlength{\belowcaptionskip}{-15pt}
\centering
\includegraphics[width=\linewidth]{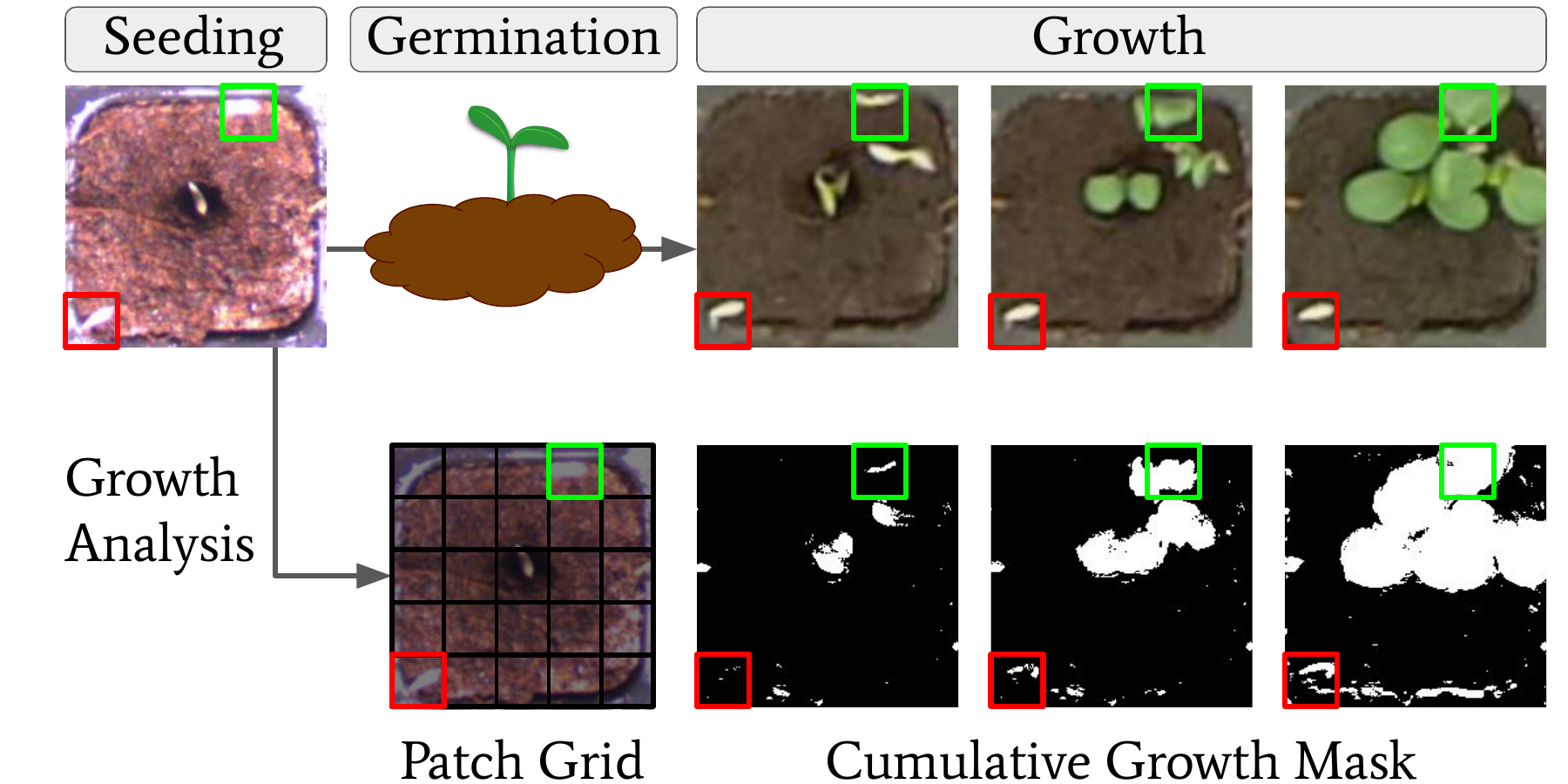}
\caption{\textbf{Plant lifecycle} with patch-based analysis of growth.}
\label{fig:patch_analysis}%
\end{figure} 

\textbf{Plant Growth:} We measure plant growth over time with a Mixture of Gaussians background subtraction module (\textbf{MOG2}~\cite{1333992,ZIVKOVIC2006773}).  
As shown in fig.~\ref{fig:patch_analysis}, this module produces a sequence of binary masks for the foreground pixels in an input sequence of images: $\mathbf{I}^t = \text{\textbf{MOG2}}(\mathbf{\bar{P}}^{t}, \mathbf{\bar{P}}^{t-1}, \mathbf{I}^{t-1})$.
We then compute a cumulative mask by accumulating each foreground increment: $C(t) = \sum{\{\mathbf{I}^0,...\mathbf{I}^t\}}$.  Our growth metric is a spatial average of the binary cumulative mask at each time step: $h(t) = \text{avg}_{[x,y]}(C(t))$.

\textbf{Seed Location vs. Plant Growth:} 
To evaluate the effect of seed location on plant growth, we subdivide each image patch into a $5 \times 5$ grid of subpatches, as shown in fig.~\ref{fig:patch_analysis}.  For each seed detected in one of these subpatches, we compute the corresponding $h_{x,y}(t)$ for that subpatch.  Finally, we compute a \textbf{growth score} $\bar{h}$ for each seed: 
\begin{equation}\label{eq:growth_score}
    \bar{h}_{x,y} = \text{avg}_t(h_{x,y}(t))
\end{equation}
\section{Results}

\subsection{Dataset}
Without loss of generality, we present results for $1$ seeding robot and $100$ growing robots, a representative subset of the full robotic system.  Our testing dataset consists of $17$ unique seeding rafts, each with a unique match among $473$ total growing rafts.  Our training dataset includes $38$ separate positive pairs.  Each of the growing rafts look visually similar, with only slight natural variations.  The seeding raft looks substantially different from the growing rafts.  These observations cover $10$ plant types and span several months.

\subsection{Multi-Robot Dense Matching}
\label{sec:dense_matching}

\textbf{Fully End-to-End Correspondence:} 
We first attempted this problem with direct raft-to-raft image matching.  Our initial attempts, as well as several state-of-the-art baselines~\cite{lindenberger2023lightglue,truong2021pdc}, were not successful.  These failures likely arose because our object of interest has (a) weak features that sometimes shift and change over time (i.e. plants) and (b) strong features that are ambiguously tessellated (i.e. raft gridding).  Our difficult agricultural setting requires methods to focus on weak features and ignore strong, ambiguous features, contrary to traditional methods.

\textbf{Alignment:}
We evaluate keypoint detections for our alignment algorithm.  The \textbf{Raft BBox NN}, \textbf{Seeding Raft Vertex NN}, \textbf{Growing Raft Vertex NN} achieve $\mathbf{4}$ pixel MSE and $\mathbf{97.9\%}$, $\mathbf{96.1\%}$, and $\mathbf{82.6\%}$ detection accuracy, respectively.  The resolution difference between the seeding and growing cameras likely accounts for the drop in performance.

\textbf{Discrete Matching:}
We show the results of our algorithm on retrieving a correct match between a query seeding raft and several candidate growing rafts, including:
\begin{itemize}
    \item One positive and one negative match (\textbf{1-vs-1})
    \item One positive and many negative matches (\textbf{1-vs-Many})
\end{itemize}
For (\textbf{Multi-Pass 1-vs-Many}), we average the computed distance between query and all candidates over $10$ different passes.
We report how frequently we place the correct match in the top 1 and 3 \textit{lowest} distances, as averaged across our $17$ matching pairs. First, we present results for a \textbf{hard negative} split of the dataset, involving \textbf{10 robots}.  Later, we show how our method generalizes to a \textbf{100 robot} dataset.

\begin{table}
\vspace{5mm}
\resizebox{\linewidth}{!}{%
\begin{tabular}{cc||c|cc|cc}
& & \multicolumn{5}{c}{Retrieval Accuracy}\\
\hline
 & & 1-vs-1 & \multicolumn{4}{c}{{1-vs-Many}}\\
 & & & \multicolumn{2}{c}{Single} & \multicolumn{2}{c}{Multi-Pass}\\
\multicolumn{2}{c||}{Method} & & top1 & top3 & top1 & top3\\
\hline
\multicolumn{2}{c||}{Siamese~\cite{koch2015siamese}} & 61.2 & 17.1 & 37.6 & 23.5 & 41.2\\
\multicolumn{2}{c||}{Metric Learning~\cite{schroff2015cvpr}} & 68.8 & 16.5 & 27.6 & 17.6 & 29.4\\
\multicolumn{2}{c||}{DeepMIL~\cite{ilse2018deepmil}} & 81.6 & 5.3 & 17.1 & 11.8 & 23.5 \\
\multicolumn{2}{c||}{Stacked Attention*~\cite{vaswani2017attention}} & 92.5 & 73.5 & 85.9 & 82.4 & 88.2\\
\midrule
\textbf{P=22}, \textbf{F=128} & \textbf{Ours} & \textbf{99.1} & \textbf{67.1} & \textbf{90.0} & \textbf{88.2} & \textbf{100}\\
\cmidrule(lr){2-7}
& 1& 73.0 & 19.4 & 47.6 & 47.1 & 82.4\\
\textbf{P:} Num Patches & 2& 85.2 & 24.7 & 56.5 & 47.1 & 70.6\\
& 4& 91.3 & 43.5 & 74.7 & 64.7 & 94.1\\
\cmidrule(lr){2-7}
& 1 & 51.8 & 7.1 & 18.8 & 11.8 & 23.5\\
\textbf{F:} Feature Dim & 2 & 57.3 & 9.4 & 21.8 & 11.8 & 29.4\\
& 4 & 75.8 & 11.2 & 31.8 & 41.2 & 70.6\\
& 16 & 98.0 & 60.6 & 81.8 & 88.2 & 100\\
\end{tabular}
}
\caption{\textbf{Network retrieval accuracy} for correctly distinguishing between a positive and negative match (1-vs-1) as well as finding a positive match among many negatives (1-vs-Many), especially with ensemble averaging (Multi-Pass).
}
\label{tab:retrieval_combined}
\end{table}

\textbf{Hard Negative Dataset (\textit{10 Robots}):} Prior work~\cite{schroff2015cvpr} emphasizes the importance of training with hard negatives, especially to distinguish between similar-looking negative instances (our dataset contains many).  For our work, we sample patch negatives from the following categories of increasing difficulty: \textit{same raft} ($40\%$), \textit{same robot} $\mathbf{G}$ ($40\%$), and \textit{all images} ($20\%$).  Patches from the same raft and robot look the most similar and thus pose the hardest challenge.

In Tab.~\ref{tab:retrieval_combined}, we show that our method outperforms several key baselines~\cite{ilse2018deepmil,vaswani2017attention,schroff2015cvpr,koch2015siamese}.  The metric learning approaches send a concatenated set of input patches into dedicated~\cite{schroff2015cvpr} or shared~\cite{koch2015siamese} feature encoders, which are then trained via a triplet loss.  Unlike our approach, these methods do not leverage intermediate patch features and feature ensembling.  DeepMIL~\cite{ilse2018deepmil} and Stacked Attention~\cite{vaswani2017attention} do use some form of feature-based ensembling, achieving improved performance.  However, these methods use a classification loss instead of a triplet loss.  Our method achieves the strongest performance thanks to both patch-feature ensembling, intermediate feature training, and metric learning.  We show that multiple passes of our patch-ensembling method improves our \textbf{1-vs-Many} accuracy, justifying this architectural choice.  

\textbf{Multi-Pass Discrete Matching:}
In Tab.~\ref{tab:retrieval_combined}, we also provide an ablation study showing that a larger subset of patches improves our overall matching performance.  Note the poor performance for a single patch, which has a \textbf{1-vs-1} accuracy of $73.0\%$, compared to $22$ patches with $99.1\%$.  Individual patches frequently lack distinctive features, so successful methods must consider ensembles of multiple patches.  However, increasing the numbers of patches (and hence computational cost) yields diminishing accuracy improvements.  We found the optimal number of patches to be \textbf{22}.

\textbf{Bandwidth-Efficient Matching:}
In Tab.~\ref{tab:retrieval_combined}, we show the trade-off between retrieval accuracy and embedding dimension.  Often, a feature size of $16$, is sufficient to obtain a strong multi-pass accuracy.  Alternatively, a size of 128 is ideal if there are time limitations and no bandwidth limitations.  These results motivated our choice of transmission policy, which sequentially transmits $10$ feature vectors of $|f|=16$ then switches to $10$ feature vectors of $|f|=128$.  

\begin{table}
\vspace{3mm}
\resizebox{\linewidth}{!}{%
\begin{tabular}{c||cc|cc}
& Total Packet & Avg Num & \multicolumn{2}{c}{Accuracy} \\
Tx Policy & Dim & Packets & top1 (\%) & top3 (\%) \\
\hline
$\alpha = \infty$ & 1440 & 20.0 & 94.1 & 100\\
$\alpha = 2.0$ & 683 & 10.7 & 90.6 & 98.7\\
$\alpha = 1.0$ & 160 & 3.8 & 83.6 & 94.1\\
$\alpha = 0.5$ & 32 & 1.5 & 75.9 & 90.3\\
$\alpha = -\infty$  & 16 & 1.0 & 61.2 & 81.1\\
\end{tabular}
}
\setlength{\belowcaptionskip}{-15pt}
\caption{\textbf{Retrieval accuracy for transmission policies.}
}
\label{tab:retrieval_tx_policy}
\end{table}
In Tab.~\ref{tab:retrieval_tx_policy}, we show the impact of our bandwidth-efficient transmission policy for different bandwidth preferences ($\alpha$ from eq.~\ref{eq:metric_learning}).  A low $\alpha$ corresponds to a policy that prefers early termination of transmissions.  This policy conserves bandwidth at the expense of accuracy. Note that our approach can still achieve a $90\%$ retrieval accuracy with only $683$ FPN, a $50\%$ reduction from the full transmission policy and several orders of magnitude cheaper ($10^4$) than raw image transmission ($\approx 1 MB$ aft compression).  This reduced dimension saves both network bandwidth and computation.

\textbf{Large-Scale Dataset (\textit{100 Robots}):} We report that our retrieval method generalizes well to large scales, achieving a pairwise matching accuracy of $\mathbf{99.8\%}$.  Moreover, we attain \textbf{top1}, \textbf{top3}, \textbf{top5}  retrieval accuracies of $\mathbf{64.7\%}$, $\mathbf{82.4\%}$, $\mathbf{100\%}$, respectively.  
On average, our choice is in the top $\mathbf{0.16}^{\text{th}}$ percentile of distances.  Despite training on a much smaller dataset, our method excels at finding a matching raft in a much larger pool of $475$ ambiguously similar rafts. This generalization was enabled by our novel patch extraction scheme (which expanded our dataset) and choice of triplet loss objective with hard-negative sampling.

\begin{table}
\vspace{3mm}
\setlength{\belowcaptionskip}{-10pt}
\resizebox{\linewidth}{!}{%
\begin{tabular}{ll|cc}
& & \multicolumn{1}{c}{Strong CPU\tablefootnote{Intel(R) Core(TM) i7-8700K CPU @ 3.70GHz}}(s) & \multicolumn{1}{c}{Weak CPU\tablefootnote{Cortex A-72 (ARM v8) 64-bit SoC @ 1.8GHz}}(s) \\
\midrule
\midrule
Align & Undistort & 0.6 & 3.0 \\ 
& BBox NN & 0.8 & 9.9 \\
& Vertex NN $\mathbf{G}$ & 11.2 & 41.6 \\
\cmidrule(lr){2-4}
Match & Patch Extraction & 0.2 & 2.6 \\
& Matching NN & 0.2 & 6.0 \\
\cmidrule(lr){2-4}
Analysis & BG Subtraction & 0.2 & 0.9 \\
\midrule
& \textbf{Total} & \textbf{13.1} & \textbf{64.0} \\
\midrule
\midrule
\end{tabular}
}
\caption{\textbf{Median processing times.}}
\label{tab:processing_time}
\end{table}
\textbf{Heterogeneous, Decentralized Compute:}
In Tab.~\ref{tab:processing_time}, we present a timing study for each stage of the matching pipeline.  With a centralized policy, 100 growing robots transmit their images to a strong centralized computer (consuming $\mathbf{\sim100}$ \textbf{MB} of bandwidth), which performs the matching task in $\mathbf{\sim21}$ \textbf{minutes}.  However, our parallelized transmission policy accomplishes the task in $\mathbf{64}$ \textbf{seconds} (a $\mathbf{20.5x}$ speedup!), while consuming only $\mathbf{8}$ \textbf{MB} (a $\mathbf{12.5x}$ reduction!).  The results scale well for a large-scale deployment (3000 robots), obtaining a $\mathbf{616x}$ speedup and $\mathbf{375x}$ reduction in bandwidth.

\subsection{Multi-View Seed-Growth Analysis}
The results from the preceding section allow us to monitor the same bench of plants from two specialized perspectives.

\begin{figure}
\vspace{2mm}
\setlength{\belowcaptionskip}{-15pt}
\centering
\includegraphics[width=\linewidth]{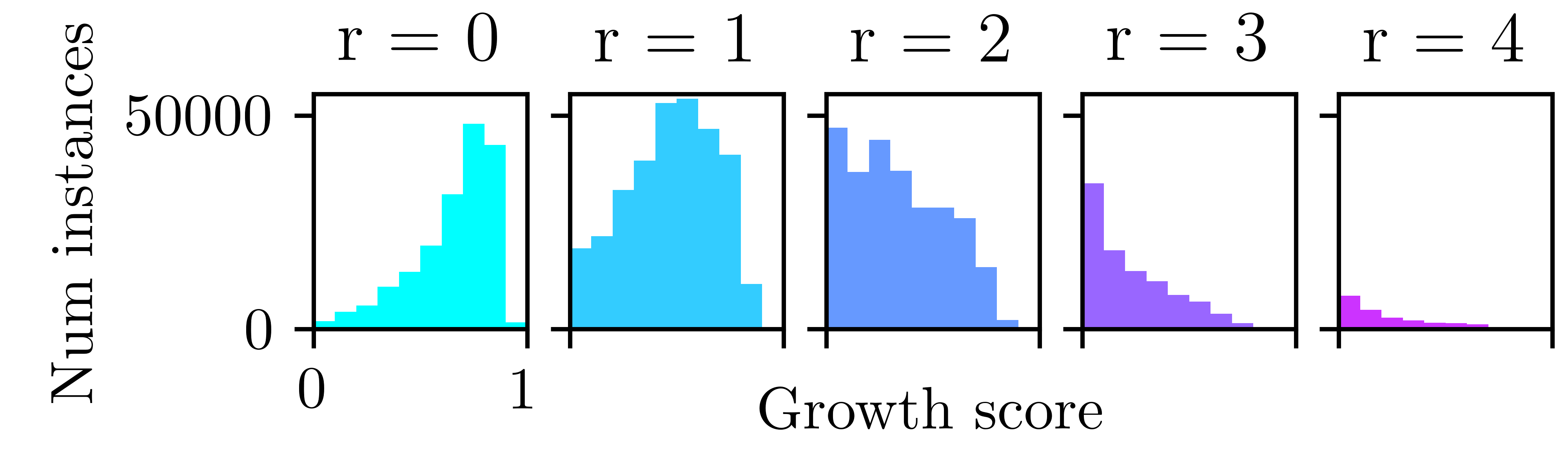}
\caption{\textbf{Histograms of growth scores} for seeds detected at increasing $l_1$ distances from the center patch ($r=0$).}
\label{fig:growth_histogram}%
\end{figure} 
\textbf{Seed Location vs. Plant Growth:}
We analyze the growth patterns of \textbf{712,636} individual seeds planted across \textbf{2,885} individual seeding cells.  As shown in Fig.~\ref{fig:growth_histogram}, we observe that seeds detected towards the center of the seeding cell have higher growth scores, as computed in eq.~\ref{eq:growth_score}.
On average, seeds in the center ($r \leq 1$) attain $52.9\%$ growth, compared to seeds towards the edges ($r > 1$) with $29.1\%$.  The seeds at the edge account for $46.8\%$ of the total number of seeds but only produce $32.6\%$ of the growth.
These results confirm our hypothesis: seeds planted far from the center of the cell experience reduced average growth than those properly planted.  Moreover, a corrective action is merited since a substantial fraction of seeds fall at this distance.

\section{Conclusion}
With \textbf{N-QR}, we tackle the task of multi-robot instance correspondence within the setting of a production-scale robotic farm.  We test our approach on an unprecedented and challenging image matching dataset, full of visually similar instances with misleading features.  We use novel multi-pass patch-ensembling to achieve a \textbf{top1} retrieval accuracy of \textbf{88.2\%}, outperforming several key baselines.
On a high-volume matching task with 100 robots, we show that our transmission policy yields a retrieval accuracy of \textbf{64.7\%} (finding a single match out of $473$ rafts), \textbf{12.5x} reduction in bandwidth, and a \textbf{20.5x} speedup.  

\textbf{Future work} will explore how our approach generalizes to other settings that significantly change over time.  It will also explore how our method better enables downstream robotics tasks, such as image-based fusion, localization, and mapping.

\printbibliography

@STRING{CVPR	= "CVPR"}

@STRING{ECCV	= "ECCV"}

@STRING{ICPR 	= "ICPR"}

@String{cvpr = "Proceedings of the IEEE Conference on Computer Vision and Pattern Recognition (CVPR)"}

@String{aaai = "Proceedings of the AAAI Conference on Artificial Intelligence (AAAI)"}

@String{eccv = "Proceedings of the European Conference on Computer Vision (ECCV)"}

@String{iros = "Proceedings of the IEEE/RSJ International Conference on Intelligent Robots and Systems (IROS)"}

@String{icml = "Proceedings of the International Conference on Machine Learning (ICML)"}

@String{icpr = "Proceedings of the International Conference on Pattern Recognition (ICPR)"}

@inproceedings{vaswani2017attention,
  title={Attention is all you need},
  author={Vaswani, Ashish and Shazeer, Noam and Parmar, Niki and Uszkoreit, Jakob and Jones, Llion and Gomez, Aidan N and Kaiser, {\L}ukasz and Polosukhin, Illia},
  booktitle={Advances in neural information processing systems},
  pages={5998--6008},
  year={2017}
}

@INPROCEEDINGS{1333992,
  author={Zivkovic, Z.},
  booktitle={Proceedings of the 17th International Conference on Pattern Recognition, 2004. ICPR 2004.}, 
  title={Improved adaptive Gaussian mixture model for background subtraction}, 
  year={2004},
  volume={2},
  number={},
  pages={28-31 Vol.2},
  doi={10.1109/ICPR.2004.1333992}}

@article{ZIVKOVIC2006773,
title = {Efficient adaptive density estimation per image pixel for the task of background subtraction},
journal = {Pattern Recognition Letters},
volume = {27},
number = {7},
pages = {773-780},
year = {2006},
issn = {0167-8655},
doi = {https://doi.org/10.1016/j.patrec.2005.11.005},
url = {https://www.sciencedirect.com/science/article/pii/S0167865505003521},
author = {Zoran Zivkovic and Ferdinand {van der Heijden}}
}

@article{chen2022deep,
  title={Deep learning for instance retrieval: A survey},
  author={Chen, Wei and Liu, Yu and Wang, Weiping and Bakker, Erwin M and Georgiou, Theodoros and Fieguth, Paul and Liu, Li and Lew, Michael S},
  journal={IEEE Transactions on Pattern Analysis and Machine Intelligence},
  year={2022},
  publisher={IEEE}
}

@InProceedings{Weyand_2020_CVPR,
author = {Weyand, Tobias and Araujo, Andre and Cao, Bingyi and Sim, Jack},
title = {Google Landmarks Dataset v2 - A Large-Scale Benchmark for Instance-Level Recognition and Retrieval},
booktitle = {Proceedings of the IEEE/CVF Conference on Computer Vision and Pattern Recognition (CVPR)},
year = {2020}
}

@article{zhao2017spatial,
  title={Spatial pyramid deep hashing for large-scale image retrieval},
  author={Zhao, Wanqing and Luo, Hangzai and Peng, Jinye and Fan, Jianping},
  journal={Neurocomputing},
  volume={243},
  pages={166--173},
  year={2017},
  publisher={Elsevier}
}

@article{song2017deep,
  title={Deep region hashing for efficient large-scale instance search from images},
  author={Song, Jingkuan and He, Tao and Gao, Lianli and Xu, Xing and Shen, Heng Tao},
  journal={arXiv preprint arXiv:1701.07901},
  year={2017}
}

@inproceedings{song2018binary,
  title={Binary generative adversarial networks for image retrieval},
  author={Song, Jingkuan and He, Tao and Gao, Lianli and Xu, Xing and Hanjalic, Alan and Shen, Heng Tao},
  booktitle={Proceedings of the AAAI conference on artificial intelligence},
  volume={32},
  number={1},
  year={2018}
}

@inproceedings{muller1999efficient,
  title={Efficient access methods for content-based image retrieval with inverted files},
  author={M{\"u}ller, Henning and Squire, David McG and Mueller, Wolfgang and Pun, Thierry},
  booktitle={Multimedia Storage and Archiving Systems IV},
  volume={3846},
  pages={461--472},
  year={1999},
  organization={SPIE}
}

@article{mokhtar2022using,
  title={Using machine learning models to predict hydroponically grown lettuce yield},
  author={Mokhtar, Ali and El-Ssawy, Wessam and He, Hongming and Al-Anasari, Nadhir and Sammen, Saad Sh and Gyasi-Agyei, Yeboah and Abuarab, Mohamed},
  journal={Frontiers in Plant Science},
  volume={13},
  pages={706042},
  year={2022},
  publisher={Frontiers}
}

@article{jung2015lettuce,
author = {Jung, Dae-Hyun and Park, Soo Hyun and Han, Xiongzhe and Kim, Hak-Jin},
year = {2015},
month = {03},
pages = {89-93},
title = {Image Processing Methods for Measurement of Lettuce Fresh Weight},
volume = {40},
journal = {Journal of Biosystems Engineering},
doi = {10.5307/JBE.2015.40.1.089}
}

@article{zhang2020growth,
  title={Growth monitoring of greenhouse lettuce based on a convolutional neural network},
  author={Zhang, Lingxian and Xu, Zanyu and Xu, Dan and Ma, Juncheng and Chen, Yingyi and Fu, Zetian},
  journal={Horticulture research},
  volume={7},
  year={2020},
  publisher={Oxford Academic}
}

@article{
riera2021deep,
author = {Luis G. Riera  and Matthew E. Carroll  and Zhisheng Zhang  and Johnathon M. Shook  and Sambuddha Ghosal  and Tianshuang Gao  and Arti Singh  and Sourabh Bhattacharya  and Baskar Ganapathysubramanian  and Asheesh K. Singh  and Soumik Sarkar },
title = {Deep Multiview Image Fusion for Soybean Yield Estimation in Breeding Applications},
journal = {Plant Phenomics},
volume = {2021},
number = {},
pages = {},
year = {2021},
doi = {10.34133/2021/9846470},
URL = {https://spj.science.org/doi/abs/10.34133/2021/9846470},
eprint = {https://spj.science.org/doi/pdf/10.34133/2021/9846470}
}

@InProceedings{schroff2015cvpr,
author = {Schroff, Florian and Kalenichenko, Dmitry and Philbin, James},
title = {FaceNet: A Unified Embedding for Face Recognition and Clustering},
booktitle = {Proceedings of the IEEE Conference on Computer Vision and Pattern Recognition (CVPR)},
year = {2015}
}

@inproceedings{koch2015siamese,
  title={Siamese neural networks for one-shot image recognition},
  author={Koch, Gregory and Zemel, Richard and Salakhutdinov, Ruslan and others},
  booktitle={ICML deep learning workshop},
  volume={2},
  number={1},
  year={2015},
  organization={Lille}
}

@article{ilse2018deepmil,
  author       = {Maximilian Ilse and
                  Jakub M. Tomczak and
                  Max Welling},
  title        = {Attention-based Deep Multiple Instance Learning},
  journal      = {CoRR},
  volume       = {abs/1802.04712},
  year         = {2018},
  url          = {http://arxiv.org/abs/1802.04712},
  eprinttype    = {arXiv},
  eprint       = {1802.04712},
  bibsource    = {dblp computer science bibliography, https://dblp.org}
}

@article{li2021deep,
  title={Deep multiple instance selection},
  author={Li, Xin-Chun and Zhan, De-Chuan and Yang, Jia-Qi and Shi, Yi},
  journal={Science China Information Sciences},
  volume={64},
  pages={1--15},
  year={2021},
  publisher={Springer}
}

@INPROCEEDINGS{orb,
  author={Rublee, Ethan and Rabaud, Vincent and Konolige, Kurt and Bradski, Gary},
  booktitle={2011 International Conference on Computer Vision}, 
  title={ORB: An efficient alternative to SIFT or SURF}, 
  year={2011},
  volume={},
  number={},
  pages={2564-2571},
  doi={10.1109/ICCV.2011.6126544}
}

@misc{muja2015fast,
  title={Fast library for approximate nearest neighbors},
  author={Muja, Marius and Lowe, David G},
  year={2015}
}

@INPROCEEDINGS{lowe1999sift,
  author={Lowe, D.G.},
  booktitle={Proceedings of the Seventh IEEE International Conference on Computer Vision}, 
  title={Object recognition from local scale-invariant features}, 
  year={1999},
  volume={2},
  number={},
  pages={1150-1157 vol.2},
  doi={10.1109/ICCV.1999.790410}}

@article{fischler1981ransac,
author = {Fischler, Martin A. and Bolles, Robert C.},
title = {Random Sample Consensus: A Paradigm for Model Fitting with Applications to Image Analysis and Automated Cartography},
year = {1981},
issue_date = {June 1981},
publisher = {Association for Computing Machinery},
address = {New York, NY, USA},
volume = {24},
number = {6},
issn = {0001-0782},
url = {https://doi.org/10.1145/358669.358692},
doi = {10.1145/358669.358692},
journal = {Commun. ACM},
pages = {381–395},
numpages = {15}
}

@inproceedings{sarlin2020superglue,
  title={Superglue: Learning feature matching with graph neural networks},
  author={Sarlin, Paul-Edouard and DeTone, Daniel and Malisiewicz, Tomasz and Rabinovich, Andrew},
  booktitle={Proceedings of the IEEE/CVF conference on computer vision and pattern recognition},
  pages={4938--4947},
  year={2020}
}

@article{lindenberger2023lightglue,
  title={LightGlue: Local Feature Matching at Light Speed},
  author={Lindenberger, Philipp and Sarlin, Paul-Edouard and Pollefeys, Marc},
  journal={arXiv preprint arXiv:2306.13643},
  year={2023}
}

@article{glaser2021enhancing,
  author       = {Nathaniel Glaser and
                  Yen{-}Cheng Liu and
                  Junjiao Tian and
                  Zsolt Kira},
  title        = {Enhancing Multi-Robot Perception via Learned Data Association},
  journal      = {CoRR},
  volume       = {abs/2107.00769},
  year         = {2021},
  url          = {https://arxiv.org/abs/2107.00769},
  eprinttype    = {arXiv},
  eprint       = {2107.00769},
  bibsource    = {dblp computer science bibliography, https://dblp.org}
}

@inproceedings{glaser2021overcoming,
  title={Overcoming obstructions via bandwidth-limited multi-agent spatial handshaking},
  author={Glaser, Nathaniel and Liu, Yen-Cheng and Tian, Junjiao and Kira, Zsolt},
  booktitle={2021 IEEE/RSJ International Conference on Intelligent Robots and Systems (IROS)},
  pages={2406--2413},
  year={2021},
  organization={IEEE}
}

@article{truong2020gocor,
  title={GOCor: Bringing globally optimized correspondence volumes into your neural network},
  author={Truong, Prune and Danelljan, Martin and Gool, Luc V and Timofte, Radu},
  journal={Advances in Neural Information Processing Systems},
  volume={33},
  pages={14278--14290},
  year={2020}
}

@article{truong2021pdc,
  author       = {Prune Truong and
                  Martin Danelljan and
                  Radu Timofte and
                  Luc Van Gool},
  title        = {PDC-Net+: Enhanced Probabilistic Dense Correspondence Network},
  journal      = {CoRR},
  volume       = {abs/2109.13912},
  year         = {2021},
  url          = {https://arxiv.org/abs/2109.13912},
  eprinttype    = {arXiv},
  eprint       = {2109.13912},
  bibsource    = {dblp computer science bibliography, https://dblp.org}
}

@inproceedings{law2018cornernet,
  title={Cornernet: Detecting objects as paired keypoints},
  author={Law, Hei and Deng, Jia},
  booktitle={Proceedings of the European conference on computer vision (ECCV)},
  pages={734--750},
  year={2018}
}

@inproceedings{he2017mask,
  title={Mask r-cnn},
  author={He, Kaiming and Gkioxari, Georgia and Doll{\'a}r, Piotr and Girshick, Ross},
  booktitle={Proceedings of the IEEE international conference on computer vision},
  pages={2961--2969},
  year={2017}
}
\end{document}